\documentclass{article}
\usepackage[T1]{fontenc}
\usepackage{spconf,amsmath,amssymb,graphicx,hyperref}
\usepackage{cite}
\usepackage{booktabs}
\usepackage{multirow}

\title{Does Location Matter? A Black-Box Causal Audit Reveals\\
Model-Dependent Audio Grounding in LALM Hallucination}

\name{
\begin{tabular}{@{}c@{}}
Mohammad Khalooei$^{1,2}$\hspace{0.8em}
Mohammad Sabokrou$^{3,4}$
\end{tabular}
}

\address{
\begin{tabular}{@{}c@{}}
$^{1}$Sharif University of Technology, Iran
\hspace{0.6em}
$^{2}$Amirkabir University of Technology, Iran
\\[1pt]
$^{3}$New Uzbekistan University, Uzbekistan
\hspace{0.6em}
$^{4}$Okinawa Institute of Science and Technology, Japan
\end{tabular}
}

\newcommand{\oraclen}{608}
\newcommand{\oracleflipO}{16.8\%}
\newcommand{\oracleflipC}{10.5\%}
\newcommand{\oracleflipR}{3.1\%}
\newcommand{\oracledropO}{0.126}
\newcommand{\oracledropC}{0.079}
\newcommand{\oracledropR}{0.016}
\newcommand{\oracleiou}{0.45}
\newcommand{\oraclezero}{25\%}

\newcommand{\qwenprofileauroc}{0.743}
\newcommand{\qwenprofilebest}{0.727}

\newcommand{\afprofilelossex}{-9.9}

\newcommand{\transferqaf}{-1.4}
\newcommand{\transferafq}{-13.1}

\newcommand{\qwenlangfix}{40.7\%}
\newcommand{\qwenlangbreak}{11.1\%}
\newcommand{\qwenaadfix}{22.4\%}
\newcommand{\qwenaadbreak}{3.6\%}
\newcommand{\qwencausafixpct}{9.1\%}
\newcommand{\qwencausabreakpct}{2.5\%}
\newcommand{\afcausafixn}{4}
\newcommand{\afcausabreakn}{45}
\newcommand{\qwencombinedacc}{70.0\%}
\newcommand{\qwencombinedcv}{71.0\%}
\newcommand{\qwencombinedcvstd}{0.7}
\newcommand{\langpriorbroadn}{48}
\newcommand{\langpriorbroadpos}{42}
\newcommand{\langpriorbroadneg}{5}
\newcommand{\langpriorworstclass}{drinking/sipping}
\newcommand{\gatedacc}{71.4\%}
\newcommand{\gatedstd}{0.8}
\newcommand{\gatedwinfrac}{79\%}
\newcommand{\gatedpval}{5.2\times10^{-9}}
\newcommand{\ensemblegatedacc}{71.9\%}
\newcommand{\ensemblegatedstd}{0.7}
\newcommand{\ensemblewinfrac}{86\%}
\newcommand{\ensemblepval}{3.6\times10^{-12}}
\newcommand{\ensembletotal}{+14.8}

\begin{document}
\ninept
\setlength{\textfloatsep}{4pt plus 1pt minus 2pt}
\setlength{\floatsep}{3pt plus 1pt minus 1pt}
\setlength{\intextsep}{4pt plus 1pt minus 2pt}
\setlength{\abovecaptionskip}{2pt}
\setlength{\belowcaptionskip}{1pt}
\renewcommand{\arraystretch}{0.8}
\maketitle

\begin{abstract}
Large audio-language models (LALMs) frequently \emph{hallucinate}
sounds that are not present. Audio-Aware Decoding (AAD)
\cite{hsu2025aad} contrasts logits under the real recording against an
all-zero ``blank'' waveform, which is out-of-distribution and conflates
``evidence absent'' with ``input malformed.'' We introduce a black-box
causal test: localize the region a claim would rely on with CLAP,
remove only that region via spectral gating, and compare against an
equally-sized random window, on three LALMs. On Qwen2-Audio-7B
($n{=}4355$), localized ablation's raw AUROC trails blank's (0.613 vs.\
0.721); we show this is a \emph{duration} confound: a duration-only
regression predicts blank's mean effect almost exactly ($0.067$ vs.\
$0.071$), and holding duration fixed, localized ablation beats random
removal at every grid cell, on both models. On Audio Flamingo 3 (AF3)
it flips $62.5\%$ of true positives vs.\ $41.5\%$ for random, and a
synthetic oracle control disentangles model dependence from CLAP
localization error. At decoding time, a language-prior contrast lifts
Qwen2-Audio's accuracy from $57.1\%$ to $68.2\%$, beating AAD, and to
$71.9\%$ (cross-validated) once combined and gated. A unified causal
profile does \emph{not} transfer between models; Qwen2.5-Omni-7B
replicates Qwen2-Audio's pattern, not AF3's. Code:
\url{github.com/khalooei/causa}.
\end{abstract}


\section{Introduction}

Large audio-language models (LALMs) such as Qwen2-Audio
\cite{chu2023qwen2audio} and SALMONN \cite{tang2024salmonn} answer free-form questions about
audio content but frequently exhibit \emph{object hallucination}:
confidently claiming a sound event is present when it is not
\cite{kuan2024understanding,kuan2025can}. The current inference-time
mitigation, Audio-Aware Decoding (AAD) \cite{hsu2025aad}, contrasts the
model's logits with real audio against logits from an all-zero
``blank'' waveform, almost certainly out-of-distribution for the
audio encoder: the contrast conflates ``evidence absent'' with ``input
malformed.''

We address this limitation with a more surgical intervention: localize
the spectro-temporal region a claim's evidence would occupy with
sliding-window CLAP (Contrastive Language-Audio Pretraining) similarity,
then physically remove \emph{only} that
region via spectral gating, leaving the rest of the real, in-distribution
recording untouched. This is a \emph{necessity} test in the sense of
Pearl's Probability of Necessity \cite{pearl2000causality,tian2000probabilities}:
if a claim survives removal of its own best-matching evidence, it was not
grounded in that evidence. We complement it with a \emph{sufficiency}
test (inserting a real exemplar into a genuine negative-control clip)
that validates the pipeline's sensitivity.

Unlike concurrent causal work that needs gradient or
internal-activation access
\cite{guo2026copo,causaltracing2026,tracinggrounding2026,frameleveltool2026},
our test is fully \emph{black-box} and training-free. On
Qwen2-Audio-7B-Instruct, localized ablation's raw AUROC trails blank's;
we show this is a duration confound that reverses once duration is
held fixed (Section~\ref{sec:durationconfound}). On NVIDIA Audio
Flamingo 3 \cite{goel2025audioflamingo3} and Qwen2.5-Omni-7B
\cite{qwen25omni2025}, raw detection value remains model-dependent.

\section{Related Work}

\textbf{Object hallucination in LALMs.} Kuan et al.\
\cite{kuan2024understanding} introduce the object-existence
discriminative-QA benchmark used here; a follow-up \cite{kuan2025can}
broadens this to temporal-order/attribute hallucinations. AAD
\cite{hsu2025aad} is the closest prior mitigation, our baseline.

\textbf{Causal sufficiency/necessity, and white-box audio grounding.}
COPO \cite{guo2026copo} trains a causal-completeness reward via RL on
image patches, needing training access; concurrent audio work traces
causal effects layer-by-layer \cite{causaltracing2026}, reads
activations under silence/unrelated-audio substitution
\cite{tracinggrounding2026}, or trains a frame-level head
\cite{frameleveltool2026}. Our tests need no training and intervene on
raw audio only.

\textbf{Content-free calibration and adaptive contrastive decoding.}
Zhao et al.\ \cite{zhao2021calibrate} calibrate against a content-free
input; M3ID \cite{favero2024m3id} contrasts an image-absent baseline,
structurally identical to our Section~\ref{sec:langprior} analysis
(adapted to audio, not novel). VCD
\cite{leng2024vcd}/ICD \cite{icd2024} contrast a \emph{distorted}
input, closer to our localized/random ablation. Concurrently, TCD
\cite{li2026tcd} gates a temporally-blurred contrast by
uncertainty/audio-reliance and TAD \cite{chang2026tad} gates a silence
contrast by token-level confidence; both gate \emph{when} to apply one
fixed contrast, where Section~\ref{sec:cdresults} instead scales
\emph{how much} of a language-prior contrast to apply.

\section{Method}

\subsection{Necessity: localized spectral-gate ablation}
For a claim ``yes'' to ``is there a sound of $o$ in the audio?'', we
slide a window (2\,s or 4\,s, 0.5\,s hop) over the clip and score each
against ``a sound of $o$'' using LAION-CLAP \cite{wu2023laionclap},
taking the highest-similarity window as $R(o)$. We attenuate energy
within $R(o)$ by $-40$ or $-100$\,dB via Short-Time Fourier Transform
(STFT) masking with a 50\,ms raised-cosine edge fade, leaving the rest of the waveform untouched, and
re-query the model (e.g.\ a true-positive ``laughter'' claim: gating
the CLAP-localized region flips it, $P(\text{yes})$: $0.905\to0.133$,
while an equal-size random window does not).

\subsection{Controls}
\textbf{Random-window.} Same ablation, same size/depth, on a uniformly
random window instead of CLAP's choice. \textbf{Blank-audio (AAD's
baseline \cite{hsu2025aad}).} An all-zero waveform, scored identically.
\textbf{Language-only.}\label{sec:langprior} No audio content block at
all (not silence), giving the object-conditioned ``yes'' prior
independent of any clip. \textbf{Real-unrelated-audio substitution.}
Following \cite{tracinggrounding2026}, the clip is replaced entirely
with a different real, verified-negative recording, reproduced here
fully black-box.

\textbf{Sufficiency.} For ``no'' claims on a negative-control clip, we
mix a real exemplar of the target class (CLAP-localized from a
separate positive clip) in at matched loudness and re-query, checking
whether the claim flips. \textbf{Scoring.} Every condition is scored
identically: a single forward pass yields next-token log-probabilities
for ``yes''/``no'', renormalized over the forced choice to give
$P(\text{yes})$, deterministic, with no sampling.

\subsection{Oracle-localization control}\label{sec:oraclemethod}
The grid above ablates only CLAP's \emph{guess} at the evidence
location, so a null result is consistent with two mechanisms: the
claim does not depend on any specific region, or CLAP's guess is wrong
enough that ablating it resembles ablating random. We disentangle
these by inserting a real
exemplar into a negative-control clip at a random but \emph{recorded}
offset, keeping only mixtures the model affirms, and ablating three
same-size windows on the same mixture: the true insertion window
(\textbf{oracle}), CLAP's own top pick (\textbf{clap}), and a uniform
random window (\textbf{random}), all at $-100$\,dB, also reporting the
temporal Intersection-over-Union (IoU) between CLAP's window and the
true one: a pure localization-accuracy number real clips cannot
provide.

\subsection{From detection to mitigation: contrastive decoding}\label{sec:cd}
Contrastive decoding \cite{li2023contrastive} (and its instantiations
AAD \cite{hsu2025aad} and VCD/ICD/M3ID
\cite{leng2024vcd,icd2024,favero2024m3id}) uses a contrast to
\emph{correct} the answer at decoding time:
\begin{equation}
d_{\text{adj}} = (1+\alpha)\,\mathrm{logit}(p_{\text{yes}}^{\text{real}})
- \alpha\,\mathrm{logit}(p_{\text{yes}}^{\text{contrast}})
\end{equation}
with $\mathrm{logit}(p)=\log\frac{p}{1-p}$, answer ``yes'' iff
$d_{\text{adj}}>0$. Reusing already-computed logits, we evaluate three
contrasts: \textbf{AAD-CD} (blank audio), \textbf{LangPrior-CD}
(Section~\ref{sec:langprior}, M3ID-style), \textbf{CAUSA-CD} (the grid's
strongest localized cell), sweeping $\alpha\in[0,3]$ and reporting the
literature-standard $\alpha{=}1$ to avoid cherry-picking. Our best
rule (Section~\ref{sec:cdresults}) needs two forward passes per query
(real, blank); the language prior is a precomputed per-\emph{class}
constant.

\subsection{Theoretical framing: from a binary $PN$ to a graded curve}\label{sec:theory}
The necessity-flip rate over true positives estimates Pearl's
Probability of Necessity, $PN = P(C(\mathrm{do}(R(o){:=}\emptyset))=\text{no}
\mid C(X)=\text{yes}, o \text{ present})$; the sufficiency-flip rate over
correct negatives estimates Pearl's Probability of Sufficiency, $PS$
\cite{pearl2000causality,tian2000probabilities}.
Unlike purely observational $PN/PS$ (generally unidentifiable without
strong assumptions \cite{tian2000probabilities}), intervening on the
same physical unit is closer to true counterfactual evidence under
monotonicity: removing genuine evidence cannot newly cause ``yes,'' nor
can inserting it decrease detection.

Ranking a clip's windows by CLAP similarity $w_{(1)},\dots,w_{(K)}$ and
nesting the intervention $\mathrm{do}(R_r)$ (remove the top $\lceil
rK\rceil$ windows, $r\in[0,1]$), monotonicity makes the flip event itself
monotone in $r$, licensing
\begin{equation}
R^\ast = \inf\{r \in [0,1] : C(\mathrm{do}(R_r)) = \text{no}\},
\end{equation}
the minimal necessary removal fraction (censored at 1 if never flipped).
This turns binary $PN$ into $PN(r):=P(R^\ast\le r\mid\cdot)$, its CDF,
with $PN(1)$ the classical case. Since $R^\ast\in[0,1]$,
\begin{equation}
\mathbb{E}[R^\ast] = \int_0^1 (1-PN(r))\,dr =: \mathrm{NAUC},
\end{equation}
the Normalized AUC, giving the deletion metric
\cite{petsiuk2018rise,fong2019extremal} an exact causal reading as the
mean of a latent counterfactual. Searching windows in \emph{random}
order gives $R^\ast_{\text{rand}}$; paired per clip, the
\emph{localization efficiency}
$\Delta_{\text{loc}}=\mathbb{E}[R^\ast_{\text{rand}}-R^\ast_{\text{clap}}]$
tests ``does
localization help'' more powerfully than independent grid cells.

\textbf{Estimation and monotonicity.} We use 6 checkpoints (1, 2, 4,
7, 11, 17 cumulative windows, $\approx\!6\%$ to full coverage of
a 10\,s clip) at $-100$\,dB, stopping at the first flip (biasing $\hat
R^\ast$ conservatively upward). Raw $P(\text{yes})$ is \emph{not}
monotone (it locally increases for $90\%$/$54\%$ of Qwen-Audio/AF3
claims), but $R^\ast$ needs only the \emph{thresholded decision} never
to revert to ``yes'', which holds almost exactly: $0.02\%$ reversals
for Qwen-Audio (CLAP order), $0\%$ elsewhere.

\subsection{A causal grounding profile, and its identifiability}\label{sec:profile}
Every quantity above answers a different question about the same claim:
$\mathrm{blank\text{-}drop}$ asks whether \emph{any} audio matters;
$\Delta$ (Section~\ref{sec:langprior}) asks whether audio matters
\emph{beyond the language prior}; the matched-control contrast asks
whether the \emph{located} region matters more than an equally-sized
\emph{arbitrary} one. We define the last as a specificity statistic,
$\mathrm{CS}_i = \mathrm{logit}(p^{\text{rand}}_i) -
\mathrm{logit}(p^{\text{loc}}_i)$, averaged over the four grid cells, and
its graded analogue $\Delta_{\text{loc}}$ (Section~\ref{sec:theory}), and
collect all five signals into a per-claim profile $z_i =
[\mathrm{blank\text{-}drop}_i,\,\mathrm{clap\text{-}sim}_i,\,\Delta_i,\,
\mathrm{CS}_i,\,\Delta_{\text{loc},i}]$.

\textbf{What this can and cannot identify.} Waveform deletion implements
$\mathrm{do}(E_o{:=}0)$ only if the removed span is the \emph{only}
causal channel disturbed. Under a \emph{window-exchangeability null}
(the CLAP span carries no more claim-relevant information than a
size-matched random span), $\mathbb{E}[\mathrm{CS}_i]=0$; a nonzero
$\hat{\mathrm{CS}}$ rejects it but does not prove identification (CLAP
bias toward salient spans would also give $\mathrm{CS}>0$), hence
Section~\ref{sec:oracle}'s synthetic control with known true location.

\section{Experimental Setup}

\textbf{Models.} Qwen2-Audio-7B-Instruct \cite{chu2023qwen2audio} is
the primary subject; NVIDIA Audio Flamingo 3
\cite{goel2025audioflamingo3} (Whisper-style encoder, Qwen2.5-7B
decoder) and Qwen2.5-Omni-7B \cite{qwen25omni2025} (Qwen2-Audio's
lineage, newer/larger) are cross-model checks. All three queried
identically black-box.

\textbf{Data.} The object-existence split of
\cite{kuan2025can}\footnote{\raggedright HF: kuanhuggingface/Audio-Hallucination\_Object-Existence\_AudioCaps-ESC50-VocalSound.},
10{,}800 rows, 53 classes from AudioCaps/ESC-50/VocalSound. Qwen2-Audio
($n{=}5000$, seed 42) yields 4355 affirmative claims (2488 correct,
1867 hallucinated, $43\%$ overclaim rate). AF3 affirms far less
(needed an explicit ``answer only Yes or No'' suffix; even so
$11.0\%$ vs.\ Qwen2-Audio's $87.1\%$), so we draw the full $n{=}10{,}800$,
yielding 1192 affirmative claims (1109 correct, 83 hallucinated,
$93.0\%$ precision). Qwen2.5-Omni ($n{=}2000$) yields 1774 affirmative
claims (1022 correct, 752 hallucinated). The oracle control
(\S\ref{sec:oraclemethod}) reuses the same benchmark's negative-control
rows and exemplar bank; no new data source.

\textbf{Metrics.} AUROC/AUPRC for hallucinated-vs-correct separation;
empirical $\hat{PN}$/$\hat{PS}$; 2000-resample bootstrap 95\%
confidence intervals (CIs).
AF3's small hallucinated count is well-powered at full-benchmark scale
($n{=}83$ hallucinated, $1109$ true positives) but its per-cell AUROC
ranges widely ($0.37$--$0.58$); flip-rate is the reliable metric for
this model.

\section{Results}\label{sec:results}

\subsection{Blank's edge is a duration confound}\label{sec:durationconfound}

\begin{table}[t]
\centering
\footnotesize
\caption{Hallucination-detection AUROC, $n=4355$ (1867 hallucinated, 2488 correct).}
\label{tab:main}
\begin{tabular}{@{}lc@{}}
\toprule
Method & AUROC \\
\midrule
CAUSA causal grounding ensemble (5-feature, \S\ref{sec:profile}) & \textbf{0.743} \\
AAD blank-audio ablation (baseline) \cite{hsu2025aad} & 0.721 \\
CAUSA localized, 4\,s / $-100$\,dB (best single cell) & 0.613 \\
CAUSA random-window control, 4\,s / $-100$\,dB & 0.577 \\
CAUSA localized, 2\,s / $-40$\,dB (original config) & 0.563 \\
CAUSA random-window control, 2\,s / $-40$\,dB & 0.550 \\
\bottomrule
\end{tabular}
\end{table}

Table~\ref{tab:main}: every \emph{single} localized cell trails
blank-audio; its edge over random-window is inconsistent ($-1.0$ to
$+3.6$ AUROC points, significant only at $-100$\,dB depth). Localized
ablation flips only $\hat{PN}=1.0\%$ of true positives vs.\ $10.8\%$
for blanking, an $\sim\!11\times$ gap, though blanking is the
out-of-distribution intervention we set out to improve on. Window size
and depth both push AUROC monotonically toward blank's, consistent
with \emph{how much} is removed dominating \emph{where}. Sufficiency
works as designed ($\hat{PS}=0.731$): the weak link \emph{appears} to
be localized removal alone.

We test the duration explanation directly: the same mask-size
confound vision-XAI's deletion metrics are built to integrate away
rather than report at one size \cite{petsiuk2018rise,fong2019extremal}
(\S\ref{sec:theory}), diagnosed here directly instead. Blank silences
100\% of a clip (median $10.0$\,s) vs.\ the grid's $2$--$4$\,s, an
unmatched comparison by construction. Fitting $P(\text{yes})$-drop as a linear
function of duration-fraction-removed, using only random-window cells
(zero localization signal by construction), predicts blank's mean drop
almost exactly: $0.067$ vs.\ $0.071$ actual on Qwen2-Audio ($n{=}4355$,
$R{=}0.21$, $p{=}2\times10^{-210}$); $0.710$ vs.\ $0.734$ on AF3
($n{=}1192$, $R{=}0.73$, $p{<}10^{-300}$). Blank sits almost exactly
on the duration-only line, so its AUROC edge is largely a duration
artifact, not evidence localization fails. Holding duration fixed
instead (localized vs.\ random-window removal of \emph{identical}
size, on the \emph{identical} clip, paired), CLAP localization wins at
every grid cell, on both models, overwhelmingly: mean drop $0.041$
vs.\ $0.016$ at $4$\,s/$-100$\,dB on Qwen2-Audio (Wilcoxon
$p{=}4.8\times10^{-46}$); $0.393$ vs.\ $0.233$ on AF3
($p{=}5.7\times10^{-43}$). Pooled over all four cells, localized
ablation's residual \emph{above} the duration-only line is large while
random-window's is $\approx\!0$ by construction: $+0.014$ vs.\
$-0.001$ on Qwen2-Audio ($t{=}19.9$, $p{=}1.4\times10^{-87}$,
$n{=}17{,}420$ paired cells), $+0.096$ vs.\ $-0.006$ on AF3
($t{=}23.1$, $p{=}2.2\times10^{-112}$, $n{=}4768$). Localization does
real, statistically dominant work once amount-removed is controlled
for; Table~\ref{tab:main}'s raw ranking measures a duration confound,
not localization validity.

\subsection{Graded necessity: thresholding costs the detector}\label{sec:gradedresults}

The adaptive search (\S\ref{sec:theory}) on all 4355 Qwen2-Audio
examples confirms the localization theory: paired localization
efficiency is real, $\widehat{\Delta}_{\text{loc}} = 0.028$ (Wilcoxon
$p{=}4.1\times10^{-27}$; CI $[0.023, 0.033]$): CLAP's ranking reliably
finds sufficient evidence with less removal than a blind search. This
hides heterogeneity: splitting classes into acoustically
\emph{transient} vs.\ \emph{continuous/ambient}, $\Delta_{\text{loc}}$
is several times larger for transient ($0.116$ vs.\ $0.017$,
Mann--Whitney $p{=}3.1\times10^{-15}$), independently reproduced in
the oracle control below, and Table~\ref{tab:main}'s headline gap
splits the same way, narrower on transient classes ($+8.1$ AUROC
points) than continuous ($+16.8$): a real, substantial, but partial
closing of the gap.

$\hat{R}^\ast$ itself is, surprisingly, a chance-level detector
(AUROC$=0.491$): $89\%$ of true positives/hallucinations never flip
even after near-total coverage, so the binary statistic discards the
graded confidence change along the way. The \emph{continuous}
$P(\text{yes})$ drop at the final checkpoint recovers it
(AUROC$=0.651$): a clean causal quantity and a good detector are not
automatically the same once censored. On AF3, efficiency is an order
of magnitude larger ($\widehat{\Delta}_{\text{loc}}{=}0.195$), censoring
nearly eliminated, yet $\hat{R}^\ast$ stays chance-level
(AUROC$=0.52$): removal amount needed and early resistance to removal
are related but distinct signals, unreliable for AF3 at any scale.

\subsection{Oracle-localization control: is the null a localizer artifact?}\label{sec:oracle}

\begin{table}[t]
\centering
\footnotesize
\caption{Oracle control, synthetic Qwen2-Audio positives with known evidence location ($n{=}\oraclen$, $-100$\,dB, matched size).}
\label{tab:oracle}
\begin{tabular}{@{}lccc@{}}
\toprule
& Oracle & CLAP & Random \\
\midrule
Flip rate & \oracleflipO & \oracleflipC & \oracleflipR \\
Mean $P(\text{yes})$ drop & \oracledropO & \oracledropC & \oracledropR \\
\bottomrule
\end{tabular}
\vspace{2pt}
\\ Mean CLAP--true-location IoU: \oracleiou{} (\oraclezero{} zero-overlap).
\end{table}

Table~\ref{tab:oracle} answers the question Section~\ref{sec:durationconfound}
leaves open: is Qwen2-Audio's near-flat gap because CLAP finds the
wrong region, or because the model does not depend on any specific
region even when known exactly? Both. CLAP's guess only moderately
overlaps the true region (mean IoU \oracleiou; \oraclezero\ zero
overlap), and ablating the true location beats ablating CLAP's guess by
a wide, significant margin (\oracleflipO\ vs.\ \oracleflipC, CI $[4.1,
8.4]$), which beats random (\oracleflipC\ vs.\ \oracleflipR, CI $[4.9,
9.9]$): a real share of the near-null result is a localizer-accuracy
artifact. But even the \emph{true} location flips only \oracleflipO\ of
these claims, far below blank's near-total effect: genuine, partial
location-insensitivity survives this control too.

\subsection{Language-prior correction: ensemble significant, alone it isn't}\label{sec:langpriorresults}

Querying each class with no audio block gives mean language-only
$P(\text{yes})=0.935$ for true positives, $0.928$ for
hallucinations, nearly identical (AUROC$=0.490$, chance): the model
leans ``yes'' $\approx\!93\%$ of the time regardless of object.
Correcting for this, $\Delta=P(\text{yes}\mid\text{real})-P(\text{yes}\mid\text{no
audio})$ gives AUROC$=0.727$; a 3-feature ensemble (blank-drop, CLAP,
$\Delta$; 5-fold cross-validation (CV)) reaches $0.743$. $\Delta$ alone
doesn't beat blank-audio (CI $[-0.009,0.020]$); the \emph{ensemble}
does per-example ($+0.022$, CI $[0.011,0.033]$) but not per-class (CI
$[-0.043,0.046]$). Adding substitution \cite{tracinggrounding2026}
as a fourth feature does not help (AUROC$=0.646$ alone; 4-feature
ensemble statistically flat at $0.743$), redundant rather than
complementary; the full 5-feature profile (Table~\ref{tab:main}) plateaus
at the same $0.743$.

\subsection{Contrastive decoding: a real accuracy gain}\label{sec:cdresults}

\begin{figure}[t]
\centering
\footnotesize
\includegraphics[width=0.88\linewidth]{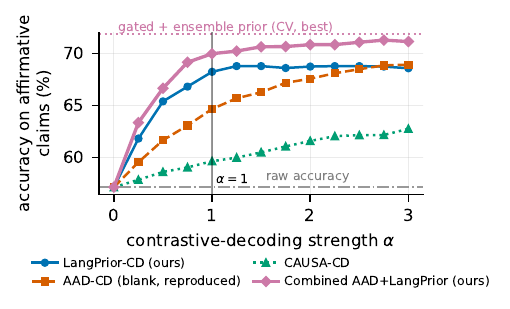}
\caption{Qwen2-Audio contrastive-decoding accuracy vs.\ $\alpha$: combining AAD+LangPrior (ours) sits above every single contrast across the whole range, and confidence-gating it (dotted line, CV) higher still (Audio Flamingo 3's sweep is flat at a $93.0\%$ ceiling for every condition and is omitted; see text).}
\label{fig:cdsweep}
\end{figure}

\begin{table}[t]
\centering
\footnotesize
\caption{Contrastive-decoding accuracy at $\alpha{=}1$, on each model's own affirmative claims. Best is swept over the literature-standard $\alpha\in[0,3]$ (Section~\ref{sec:cd}).}
\label{tab:cd}
{\setlength{\tabcolsep}{1.5pt}
\begin{tabular}{@{}llccc@{}}
\toprule
Model (raw acc.) & Contrast & Acc.\ & $\Delta$ & Best \\
\midrule
\multirow{6}{*}{Qwen2-Audio (57.1\%)}
 & LangPrior-CD & 68.2\% & $+11.1$ & 68.8\% \\
 & AAD-CD       & 64.7\% & $+7.5$ & 68.9\% \\
 & CAUSA-CD (loc.\ only) & 59.6\% & $+2.5$ & 62.8\% \\
 & AAD+LangPrior          & 70.0\% & $+12.9$ & 71.3\% \\
 & + gated & 71.4\% & $+14.3$ & --- \\
 & \textbf{+ ensemble prior (ours)} & \textbf{71.9\%} & \textbf{+14.8} & --- \\
\midrule
\multirow{3}{*}{Audio Flamingo 3 (93.0\%)}
 & LangPrior-CD & 93.0\% & $+0.0$ & 93.0\% \\
 & AAD-CD       & 93.0\% & $+0.0$ & 93.0\% \\
 & CAUSA-CD     & 89.6\% & $-3.4$ & 93.0\% \\
\bottomrule
\end{tabular}}
\end{table}

Unlike $\Delta$'s non-significant detection AUROC, the same contrast as
a decoding correction gives a substantial, monotone gain: $+11.1$ at
$\alpha{=}1$ (Table~\ref{tab:cd}, Fig.~\ref{fig:cdsweep}), with no sign
reversal in $[0,3]$; detection and correction can disagree. AAD gives
$+7.5$; CAUSA-CD alone the smallest gain; combined and gated (below)
reaches $+14.3$. On AF3, whose language prior is near zero per class,
LangPrior-CD/AAD-CD give $+0.0$ and CAUSA-CD \emph{hurts} ($-3.4$).
Aggregate accuracy hides a fix/break trade-off: on Qwen2-Audio,
LangPrior-CD fixes $\qwenlangfix$ of hallucinations at the cost of
breaking $\qwenlangbreak$ of correct claims; AAD-CD is more
conservative ($\qwenaadfix$/$\qwenaadbreak$); CAUSA-CD fixes fewest
($\qwencausafixpct$/$\qwencausabreakpct$). On AF3, CAUSA-CD's loss is
fully explained: it fixes only $\afcausafixn$ hallucinations while
breaking $\afcausabreakn$ correct claims, $11$:$1$. The gain is
broad-based on Qwen2-Audio: of $\langpriorbroadn$ classes with
$\geq\!15$ examples, $\langpriorbroadpos$ improve, $\langpriorbroadneg$
worsen (\langpriorworstclass\ breaks twice as many as it fixes).

Since AAD-CD and LangPrior-CD correct different, only partially
overlapping errors, we combine them:
\begin{equation}
d=(1{+}\alpha_b{+}\alpha_l)\mathrm{logit}(p_{\text{real}})
-\alpha_b\,\mathrm{logit}(p_{\text{blank}})-\alpha_l\,\mathrm{logit}(p_{\text{lang}})
\end{equation}
at $\alpha_b{=}\alpha_l{=}1$: accuracy reaches $\qwencombinedacc$; its
own swept best (over $\alpha_b,\alpha_l\in[0,3]$) reaches $71.3\%$,
$+2.4$ over either single contrast's own best: substitutable, not
redundant. 5-fold CV (fit $(\alpha_b,\alpha_l)$ per
training fold) gives $\qwencombinedcv\pm\qwencombinedcvstd$pp,
\emph{confirming} the gain.

\textbf{Confidence-gating and a less-noisy prior help further.} Scaling
LangPrior's weight by $p_{\text{lang}}$ itself ($w=\alpha_l\,
p_{\text{lang}}$ replacing $\alpha_l$, as TCD \cite{li2026tcd}/TAD
\cite{chang2026tad} gate a temporal contrast) gives
$\gatedacc\pm\gatedstd$pp CV, beating flat weighting in
$\gatedwinfrac$ of 20 splits (paired $t$-test, $p{=}\gatedpval$).
Averaging 5 paraphrasings of $p_{\text{lang}}$ instead of one gives
$\ensemblegatedacc\pm\ensemblegatedstd$pp ($\ensembletotal$ over raw),
beating single-phrasing in $\ensemblewinfrac$ of folds (paired
$t$-test, $p{=}\ensemblepval$), 265 extra passes, nearly free. On AF3,
combining or gating changes nothing.

\subsection{Cross-model generalization: a third model, opposite pattern on AF3}

\begin{table}[t]
\centering
\footnotesize
\caption{Necessity flip rate, localized vs.\ random, by model ($n{=}510$/$1022$/$1109$ Qwen2-Audio/Qwen2.5-Omni/AF3 true positives).}
\label{tab:crossmodel}
{\setlength{\tabcolsep}{1.5pt}
\begin{tabular}{@{}lcccc@{}}
\toprule
Model & Setting & Localized & Random & Gap \\
\midrule
Qwen2-Audio & 4\,s/$-100$\,dB & \textbf{3.3\%} & 2.2\% & $+1.2$ \\
Qwen2.5-Omni & 4\,s/$-100$\,dB & \textbf{7.4\%} & 3.9\% & $+3.5$ \\
Audio Flamingo 3 & 4\,s/$-100$\,dB & \textbf{62.5\%} & 41.5\% & $\mathbf{+21.0}$ \\
Qwen2-Audio & 2\,s/$-40$\,dB & \textbf{1.0\%} & 0.8\% & $+0.2$ \\
Qwen2.5-Omni & 2\,s/$-40$\,dB & 1.5\% & 0.8\% & $+0.7$ \\
Audio Flamingo 3 & 2\,s/$-40$\,dB & \textbf{26.0\%} & 18.8\% & $\mathbf{+7.1}$ \\
\bottomrule
\end{tabular}}
\end{table}

The localized-vs-random gap is $7$--$21$ points for AF3 against
$0$--$4$ for both Qwen-family models. Blanking flips \emph{every} AF3
true positive vs.\ $10.8\%$ for Qwen2-Audio; AF3 affirms only $11.0\%$
of queries (vs.\ $87.1\%$) but at $93.0\%$ precision (vs.\
$57.1\%$), plausibly tied to its claims being more localizable. A
third model tests whether the Qwen2-Audio/AF3 split is a two-point
coincidence: Qwen2.5-Omni-7B ($n{=}1774$, Qwen2-Audio's lineage but
newer/larger) replicates Qwen2-Audio's pattern, not AF3's (best
localized cell AUROC $0.609$ vs.\ blank's $0.766$, $+15.6$pp, CI
$[+12.5,+18.6]$), suggesting (with only three models, a hypothesis)
that the split tracks encoder lineage more than scale. Decoding-side, its overclaiming is far less
language-prior-driven (AAD-CD/LangPrior-CD move accuracy by only
$+0.7$/$+0.3$pp), but CAUSA-CD, weakest for Qwen2-Audio, gives this
model's \emph{largest} single-contrast gain ($+1.6$pp).

\subsection{Cross-model transfer of the causal grounding profile}\label{sec:profileresults}

If grounding failure has a common causal structure, a detector
\emph{trained} on one model's profile $z$ (\S\ref{sec:profile}) should
transfer to another's. In-distribution, a 5-fold CV ensemble over $z$
only modestly beats its best single signal on Qwen-Audio
($\qwenprofileauroc$ vs.\ $\qwenprofilebest$, CI crosses zero by-class)
and \emph{underperforms} on AF3 ($\afprofilelossex$\,pp), with only 83
AF3 hallucinations, so it overfits noise. The sharper test: standardize $z$
within each model, fit the logistic router on one, score the other
\emph{without retraining}. Both directions lose to the target's own
best fixed contrast (Qwen$\to$AF3: $\transferqaf$\,pp; AF3$\to$Qwen:
$\transferafq$\,pp), a genuine negative result: the two failure modes
differ qualitatively (language-prior override vs.\ coarse localization
sensitivity).

\section{Discussion, Limitations, and Conclusion}

\textbf{Limitations.} Qwen2-Audio's weak localized-vs-random gap
(\S\ref{sec:oracle}) is only \emph{partly} a CLAP-accuracy artifact;
the language-prior ensemble's significance holds per-example, not
per-class; we do not correct for multiple comparisons. Qwen2.5-Omni
has only headline numbers; Phi-4-multimodal and FLAM \cite{wu2025flam}
remain non-functional (dependency incompatibilities). On the shorter
HalluAudio benchmark \cite{halluaudio2026} ($3.8$\,s vs.\ $10.0$\,s
median), \S\ref{sec:durationconfound}'s matched-duration effect
attenuates sharply, though a small residual survives on Qwen2-Audio
($p{=}8.9\times10^{-5}$). Combined-decoding (\S\ref{sec:cdresults})
replicates cleanly regardless ($81.2\%\pm2.4$ vs.\ $80.1\%\pm2.1$).

\textbf{Conclusion.} Localization's raw detection value is model- and
class-dependent, but its weakness on Qwen2-Audio is a duration
confound: holding duration fixed, localized removal decisively beats a
matched random window on both models (\S\ref{sec:durationconfound}).
A single causal grounding profile does not transfer across models
(\S\ref{sec:profileresults}). Together with a validated decoding-time
correction, this is our contribution.

\section*{Compliance with Ethical Standards}
This study uses only publicly released benchmark audio (AudioCaps,
ESC-50, VocalSound, via \cite{kuan2025can}) and openly released
pretrained models; it involves no new data collection from human or
animal subjects and required no institutional ethical review.

{\scriptsize
\bibliographystyle{IEEEbib}
\bibliography{refs}
}

\end{document}